# Exploring the design space of diffusion and flow models for data fusion

Niraj Chaudhari[1], Manmeet Singh[2,3], Naveen Sudharsan[3], Amit Kumar Srivastava[4], Harsh Kamath[3], Dushyant Mahajan[1], Ayan Paul[1]

[1] Northeastern University, Boston, MA, USA
[2] Western Kentucky University, Bowling Green, KY, USA
[3] The University of Texas at Austin, Austin, TX, USA
[4] Leibniz-Centre for Agricultural Landscape Research (ZALF), Berlin, Germany

Correspondence: manmeet.singh@utexas.edu, a.paul@northeastern.edu

**Abstract**
Data fusion is an essential task in various domains, enabling the integration of multi-source information to enhance data quality and insights. One key application is in satellite remote sensing, where fusing multi-sensor observations can improve spatial and temporal resolution. In this study, we explore the design space of diffusion and flow models for data fusion, focusing on the integration of Defense Meteorological Satellite Program's Operational Linescan System (DMSP-OLS) and Visible Infrared Imaging Radiometer Suite (VIIRS) nighttime lights data. Our approach leverages a diverse set of 2D image-to-image generative models, including UNET, diffusion, and flow modeling architectures. We evaluate the effectiveness of these architectures in satellite remote sensing data fusion, identifying diffusion models based on UNet as particularly adept at preserving fine-grained spatial details and generating high-fidelity fused images. We also provide guidance on the selection of noise schedulers in diffusion-based models, highlighting the trade-offs between iterative solvers for faster inference and discrete schedulers for higher-quality reconstructions. Additionally, we explore quantization techniques to optimize memory efficiency and computational cost without compromising performance. Our findings offer practical insights into selecting the most effective diffusion and flow model architectures for data fusion tasks, particularly in remote sensing applications, and provide recommendations for leveraging noise scheduling strategies to enhance fusion quality.

## 1. Introduction

Nighttime light observations from satellites have become a ubiquitous data source for monitoring urbanization, economic activity, and electrification at regional to global scales [1]. The Defense Meteorological Satellite Program – Operational Linescan System (DMSP-OLS) provided the first global night lights dataset spanning 1992–2013, which has been extensively used to map urban extent and socio-economic activity [1]. However, DMSP's data suffer from coarse spatial resolution (~1 km) and limited radiometric range (6-bit digitization, prone to saturation) [2]. In contrast, the newer VIIRS Day/Night Band (VIIRS-DNB) sensor (available since 2012) offers finer spatial detail (~500 m) and a higher dynamic range (14-bit radiance measurements), enabling detection of much dimmer lights [2]. The VIIRS data dramatically reduce issues like pixel saturation and "over-glow" blooming of bright city cores. Despite these improvements, directly combining DMSP and VIIRS observations into a long-term consistent time series is challenging because the two sensors' outputs are not directly comparable [2]. The discontinuity in 2013 – when VIIRS replaced DMSP – creates a gap that hinders studies requiring multi-decade trends (e.g. urban growth, electrification progress).

A significant body of work has attempted to cross-calibrate or harmonize DMSP and VIIRS NTL data. Traditional approaches apply statistical adjustments or regressions to make VIIRS data "DMSP-like" (or vice versa). For example, inter-calibration models have been developed to convert VIIRS radiances into DMSP-equivalent Digital Numbers. Zheng et al. (2019) used a geographically weighted regression to calibrate VIIRS and DMSP at regional scales [3]. Zhao et al. (2019) fit a sigmoid function to integrate DMSP and VIIRS in Southeast Asia [4]. Globally, Li et al. (2020) generated a harmonized dataset by first inter-calibrating DMSP across years and then using the overlap year (2013) to adjust VIIRS into a "DMSP-like" radiance domain [1]. These efforts have produced continuous NTL time series [1], but they often rely on parametric fits that may not capture fine-grained spatial details or non-linear differences between sensors. Recently, machine learning has been explored: for instance, Nechaev et al. (2021) trained a deep residual U-Net to directly translate VIIRS images into DMSP-like images [5], showing improved preservation of spatial patterns (like urban cores and dim peri-urban areas) compared to simpler calibration.

Building on these developments, our work investigates *generative models* for cross-sensor nighttime light data fusion. Generative approaches, which learn the full conditional distribution of DMSP-like imagery given a VIIRS input, have the potential to model complex relationships beyond simple one-to-one mappings. In particular, we explore state-of-the-art generative modeling frameworks: diffusion models and flow matching models. Diffusion models (derived from score-based generative modeling) have recently achieved remarkable success in image synthesis, outperforming GANs in fidelity and diversity [6]. These models progressively denoise random noise into an image, guided by a learned score function (often implemented as a U-Net) that is conditioned on the input image in our case. Flow matching models, on the other hand, learn a continuous probability flow, transforming a simple base distribution (e.g. Gaussian noise) into a complex target data distribution. Flows can be naturally adapted for conditional generation by conditioning their transformations on the input image. Both approaches provide flexible ways to generate a *distribution* of plausible DMSP-like outputs given a VIIRS image, rather than a single deterministic guess.

In this paper, we explore the design space of these generative models for the VIIRS-to-DMSP translation task. We implement a range of model variants and training strategies, including: (1) a standard diffusion probabilistic model with a convolutional U-Net backbone (similar to DDPM [8]), (2) a diffusion model trained with the Latent Consistency Model (LCM) noise scheduler [22], which defines a noise variance schedule and timestep mapping designed to enable high-quality generation with very few denoising steps, (3) an implementation of Karras et al.'s EDM (Elucidated Diffusion Model) guidelines [11] for improved sampling efficiency, and (4) flow models [12]. By comparing these diverse methods on a common dataset, we aim to elucidate trade-offs in terms of reconstruction accuracy, visual quality, spectral fidelity, and computational performance.

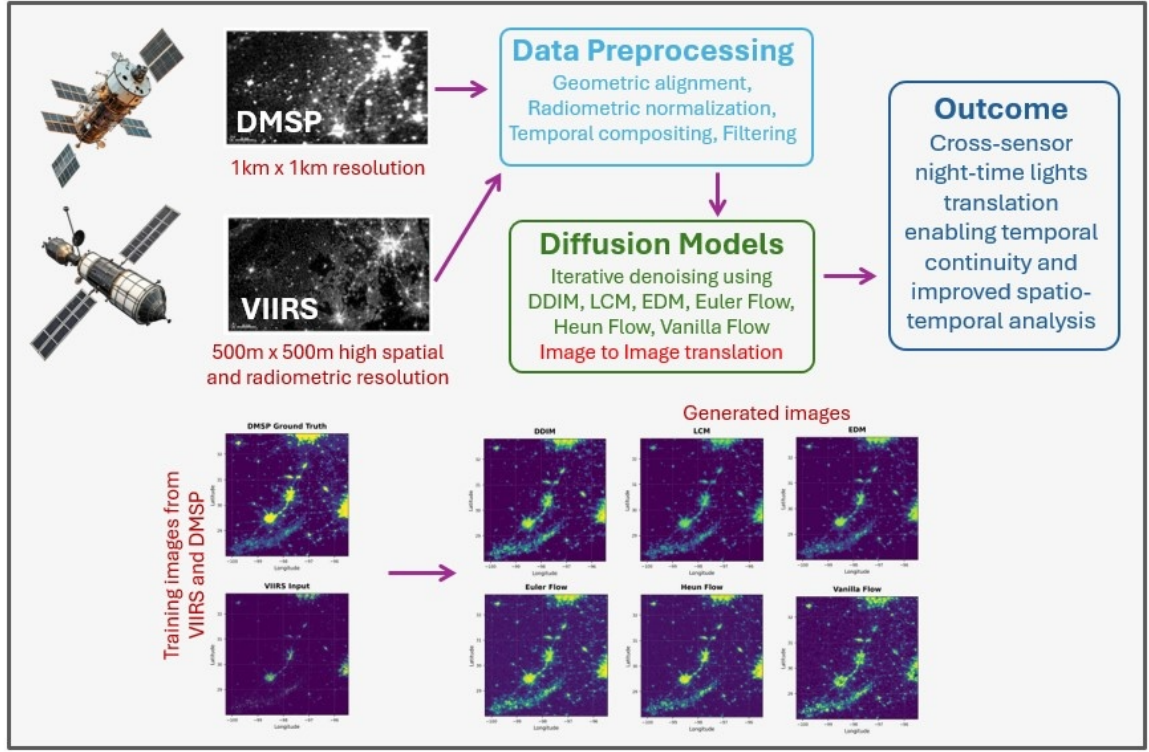

*Figure 1. Schematic of our cross-sensor data fusion pipeline. We create training pairs of VIIRS and DMSP images (after preprocessing to align spatial resolution and normalize radiometry). Generative diffusion models operating in an image-to-image translation setting (conditional denoising) are trained to produce DMSP-like outputs from VIIRS inputs. We experiment with various diffusion sampling methods: DDIM, Consistency scheduler (labeled "LCM" in results), and EDM (Karras et al.'s optimized design). We also implement flow-based models: both conventional normalizing flows (reversible networks) and diffusion-derived flows using Euler or Heun ODE solvers. Given a new VIIRS input, the models generate a synthetic DMSP image, enabling a continuous nighttime light time series across the DMSP–VIIRS transition.*

Our contributions are summarized as follows: (i) We develop a unified framework to train conditional diffusion and flow models for NTL data fusion and propose a hybrid approach leveraging diffusion-based score modeling with ODE integration for deterministic mapping. (ii) We conduct a thorough evaluation on real NTL data, using multiple image quality metrics (SSIM, PSNR, MAE) and introducing a PSD-based analysis to assess how well models replicate the scale-dependent characteristics of DMSP imagery. (iii) We analyze the computational efficiency of each approach, including the impact of half-precision (FP16) inference and different noise schedules on performance. (iv) We discuss broader implications of generative data fusion in remote sensing, such as improved long-term monitoring of urban dynamics, as well as cautionary considerations regarding the use of AI-generated geospatial data.

The rest of this paper is organized as follows. Section 2 reviews related work on nighttime lights calibration and generative models. Section 3 details our methodology, including data preprocessing and model implementations. Section 4 presents experimental results, comparing the models' accuracy and efficiency. Section 5 discusses broader impacts and limitations. Finally, Section 6 concludes the paper with potential future directions.

## 2. Related Work

**Nighttime Lights Data Fusion and Calibration:** Early studies integrating DMSP-OLS and VIIRS-DNB data highlighted the need to correct inter-sensor discrepancies. Initial empirical methods calibrated DMSP data and adjusted VIIRS values to extend the DMSP time series (Elvidge et al., 2014; Li & Zhou, 2017). Subsequent research introduced cross-sensor mapping approaches: Zheng et al. (2019) used robust regression for China, while Zhao et al. (2019) fitted VIIRS to DMSP via sigmoid functions for Southeast Asia. Data-driven methods later emerged. Li et al. (2020) harmonized global NTL data by stabilizing DMSP series and fitting grid-level polynomials between overlapping years, enabling bidirectional conversion between DMSP and VIIRS. Deep learning methods further advanced fusion: Nechaev et al. (2021) applied a Residual U-Net for image-to-image translation, reproducing DMSP's saturation and diffuse glow from VIIRS input. Building on this, our work employs generative diffusion models to capture the one-to-many mapping from high-resolution VIIRS to lower-resolution DMSP, representing uncertainty and natural variation in NTL appearance.

**Diffusion Models:** Diffusion probabilistic models (Ho et al., 2020; Song et al., 2021) define a noise-adding forward process and a learned denoising reverse process. Later improvements—DDIM (Song et al., 2021) and classifier-free guidance (Nichol & Dhariwal, 2021)—enhanced quality and efficiency. Latent diffusion (Rombach et al., 2022) further accelerated inference by operating in compressed latent space. We adopt these advances, employing a cosine noise schedule and conditional guidance. We also integrate design choices from Karras et al. (2022), using Heun and Euler ODE solvers for efficient deterministic sampling.

**Applications to Remote Sensing:** Harmonized DMSP–VIIRS data enable consistent, long-term analysis of socioeconomic and environmental dynamics. Prior studies linked NTL to electricity use, GDP, poverty, conflict, and urbanization (Zhou et al., 2018; Li & Zhou, 2017). Our fused dataset extends these applications by translating high-resolution VIIRS detail into historical periods, enhancing detection of intra-urban patterns, light pollution trends, and anomaly events such as power outages or conflict impacts. To our knowledge, this is the first use of diffusion and flow-based generative models for cross-sensor calibration of NTL imagery.

## 3. Methodology

### 3.1 Study Area and Data Preparation

Our experiments focus on a region in Texas, USA, encompassing the corridor from San Antonio and Austin up to the Dallas–Fort Worth metroplex. This region provides a diverse mix of brightly lit urban centers, smaller towns, and rural areas, making it ideal for evaluating model performance across varying light intensity levels. VIIRS and DMSP data were obtained for a common year in which both sensors operated (2013) to construct training pairs. We used the VIIRS monthly cloud-free composite images from the Suomi-NPP satellite (averaged to annual, if necessary) and the DMSP-OLS annual *stable lights* composite for 2013 (satellite F18). Both datasets were resampled to a common geographic grid covering Texas. The VIIRS radiances (original ~500 m resolution) were aggregated to ~1 km to match DMSP's footprint and to minimize resolution disparities. This was done by averaging the VIIRS pixels within each DMSP pixel area, after applying the VIIRS provided mask to remove ephemeral lights and noise (e.g., stray light, aurora effects). We also normalized the radiometric scales: DMSP stable lights are unitless digital values 0–63, while VIIRS is in radiance units (nW/cm²/sr). We linearly scaled the VIIRS radiances to 0–63 range using the overlap region statistics (essentially creating a rough initial calibration so that both inputs to the model lie in a comparable value range). This helps the learning process focus on residual differences (spatial texture, saturation effects) rather than an overall offset. After this *geometric alignment and radiometric normalization*, we applied light filtering such as removing background noise: any pixel with negligible radiance (below ~0.5 nW/cm²/sr) and no corresponding DMSP light was set to zero in both datasets to avoid confusing the models with spurious faint glows (e.g. from gas flares or background noise).

We extracted 330 non-overlapping 32×32 pixel patches from a 576×576 image for model training. Patch sampling was done to ensure a variety of scenes: we included patches centered on dense urban regions, transitional suburban areas, and sparsely lit rural regions. In total, 270 patches were used for training, with 50 reserved for validation and early stopping. Once trained, the model was applied to the entire 576×576 image, requiring it to generalize the non-linear mapping from local patches to the full scene. Prior to feeding into models, each patch's pixel values were normalized to [0,1] (dividing by 63 for DMSP and equivalent scaled range for VIIRS) and then further rescaled to [–1, 1]. We also experimented with log-transforming radiances to emphasize low-light differences but found linear normalization sufficient when the model capacity is high.

We evaluate image prediction quality with common metrics: Structural Similarity Index (SSIM), Peak Signal-to-Noise Ratio (PSNR), and Mean Absolute Error (MAE). SSIM and PSNR emphasize perceptual and pixelwise fidelity respectively, while MAE (in the original radiance units) gives an absolute calibration error measure. These are computed against the ground truth DMSP image for each patch, then averaged over the test set. In addition, we compute the Power Spectral Density (PSD) of each image and specifically use the *azimuthally averaged PSD* (i.e., radial frequency spectrum) to analyze how well the spatial frequency content matches real DMSP. This is important because DMSP images are smoother (lacking high-frequency detail) compared to VIIRS; a model that under- or over-smooths can be diagnosed via the PSD curve. We include a *zonal energy spectrum* comparison which essentially plots PSD vs spatial frequency (wavenumber).

**3.2 Model Architectures and Training**

**Diffusion Model Implementation:** Our conditional diffusion model uses a U-Net backbone to represent the noise-removal network $\epsilon_\theta(x_t, y)$, where $x_t$ is a noisy version of the target (DMSP) image at diffusion timestep $t$, and $y$ is the conditioning input (the VIIRS image). At training time, we add noise to the ground truth DMSP patch $x_0$ according to a schedule $\beta_t$ (either linear or cosine) to obtain $x_t$. The network is trained to predict the added noise $\epsilon$ given $x_t$ and $y$, using the standard simplified loss:

$$L(\theta) = E\left[\left(\epsilon - \epsilon_\theta(x_t, t, y)\right)^2\right],$$

$$x_t = \sqrt{\alpha_t}x_0 + \sqrt{1-\alpha_t}\epsilon.$$

The conditioning $y$ (VIIRS) is provided to the U-Net by channel-wise concatenation with $x_t$ at the input. This helps inject the guiding image's features. The U-Net has ~4 downsampling levels, with the highest resolution at 32×32 and lowest at 8×8 feature maps. We include self-attention blocks at 16×16 resolution to help capture mid-range dependencies (this is common in diffusion models for images ~64 or larger). We trained models on 32×32 patches with a batch size of 32. Training was conducted with a maximum limit of 1,500 epochs, but early stopping was applied, and models typically converged between 300 and 500 epochs. The cosine schedule starts with small noise increments (preserving details longer) and then larger increments later – effectively a slower diffusion at first, which seems to suit preserving the fine structures from VIIRS. During inference, we primarily use the deterministic DDIM sampling (with 30 time steps) to generate images, as it offers a good balance of speed and quality. We also test *EDM's* sampler with only 30 steps using a Heun solver (see Section 4.3). For probabilistic sampling, we can produce multiple stochastic samples per input to assess variability.

**Consistency Models**: In one experiment, we explored consistency-based generative modeling, where the objective is to learn a function that maps noisy inputs directly toward clean data in a single step, or in just a few iterative refinements. For this, we employed the LCM noise scheduler, which defines the noise variance schedule and timestep mapping in a way that supports stable training and efficient inference. A key advantage of this approach is the potential for fast sampling, since high-quality outputs can often be generated in only a handful of denoising steps, compared to the hundreds typically required by standard diffusion models. This reduction in sampling time makes consistency models an attractive alternative when computational efficiency is a priority, though we observed that matching the highest image quality of full diffusion models can still require careful tuning of the scheduler and training dynamics.

**Flow Matching Implementation**: Our "vanilla flow matching" model follows the recently proposed flow matching framework [12], which learns a continuous probability flow between Gaussian noise and target images. Unlike normalizing flows, this approach does not rely on invertible transformations or Jacobian determinants. Instead, training is framed as a regression problem: given a point along a straight-line interpolation between noise and data, the model is tasked with predicting the corresponding velocity field that transports the noisy input toward the data distribution. To incorporate conditioning, the interpolated input is concatenated with the VIIRS image, and a UNet architecture is used to predict the velocity. The model is trained by minimizing a simple L2 loss between predicted and ground-truth velocities, avoiding the computational burden of likelihood training. While inference still proceeds iteratively, flow matching models can, in principle, support fewer denoising steps compared to diffusion, offering a pathway toward faster sampling.

**Training Details and Hyperparameters:** All models were implemented in PyTorch. Training was conducted on an NVIDIA A6000 GPU (48 GB memory). The diffusion models took about an hour for ~350 epochs (with float 32 precision). Noise Schedules: We tried both linear and cosine schedules for diffusion. The linear schedule linearly increases noise variance $\beta_t$ from $\beta_1 = 1e^{-4}$ to $\beta_T = 0.02$ over $T = 1000$ steps (for training). The cosine schedule uses $\beta_t = 1 - \cos^2\big((t/T + 0.008)/(1 + 0.008)\pi/2\big)$ (Nichol & Dhariwal, 2021), which starts very small and then increases sharply near $T$. For this experiment, we used linear schedular. We did not implement advanced step-size controllers or dynamic thresholding from Karras et al. (EDM) explicitly, aside from the fixed 30-step heuristic sampler. All models were trained using an initial learning rate of $1 \times 10^{-3}$, with a cosine annealing (CosineAnnealingLR) scheduler that gradually decayed the learning rate to a minimum of $1 \times 10^{-6}$ across training epochs. The optimization objective was formulated as a mean squared error (MSE) loss under a noise prediction framework, consistent with standard diffusion model training practices. Each model was trained with a batch size of 32. To ensure generalization and prevent overfitting, early stopping was applied with a patience of 200 epochs, and the best-performing checkpoint was selected based on the lowest validation loss observed during training.

**3.3 Probability Flow ODE Sampling**

A diffusion model can be viewed as learning a *score function* $s_\theta(x_t, t, y) \approx \nabla_{x_t} \log p(x_t|y)$, which corresponds to an ODE known as the probability flow ODE (Song et al., 2021). Solving this ODE from $t = 1$ (pure noise) to $t = 0$ yields a *deterministic* image generation given a noise seed. We implemented the probability flow ODE for our conditional model using the Euler and Heun methods. The Euler method updates $x_{t-\Delta t} = x_t + f(x_t, t, y)\Delta t$ where $f$ is the ODE drift derived from the score. The Heun method is a 2nd-order Runge–Kutta that does a predictor-corrector: it first does a Euler prediction and then corrects using the average of the derivative at the beginning and end of the interval [11]. We used 30 steps for Euler and 30 for Heun in our tests (we found going below ~20 steps started to noticeably degrade quality for Euler, whereas Heun remained stable down to 20-30 steps with only slight loss). Using these ODE solvers, we produce what we term **Euler Flow** and **Heun Flow** results – these are effectively normalizing flow outputs (since the ODE integration is an invertible mapping from initial noise to image) but using the learned score/diffusion model's guidance. One advantage is that these avoid the randomness of diffusion sampling and can improve certain metrics like PSNR by eliminating sampling variance. We will compare Euler vs. Heun to demonstrate the benefit of a higher-order solver in matching the true underlying generative path (Heun tends to reduce errors like overshooting that Euler might make in one step).

**Mixed Precision and Quantization:** All training computations were performed in full precision (FP32) to ensure numerical stability and consistency across all operations, including convolutions, matrix multiplications, loss accumulation, and normalization steps. For inference, we leveraged FP16 quantization aggressively. This gave a substantial speed-up as shown in Section 4.4, especially for the diffusion models, which are compute-heavy. The flow model similarly benefited from half-precision. We did not notice any degradation in output image metrics when using FP16 inference, which is expected since the networks were trained with robust losses. We also explored 8-bit quantization for the flow model weights to potentially deploy on lower-memory environments; preliminary tests showed <0.01 decrease in SSIM with 8-bit weight quantization, but we stick to FP16 for simplicity.

**4. Results**

**4.1 Qualitative Comparison of Generated Nighttime Lights**

We first examine the *visual quality* of the fused nighttime light images produced by each method. **Figure 2** presents example outputs for a representative 64×64 pixels patch including parts of the

Austin metro area and surrounding rural regions. This figure uses a common color scale normalized to the DMSP data range (0–63 digital numbers, mapped to a color gradient from dark (purple) to bright (yellow) for visualization). The ground truth DMSP and the input VIIRS patch are shown for reference.

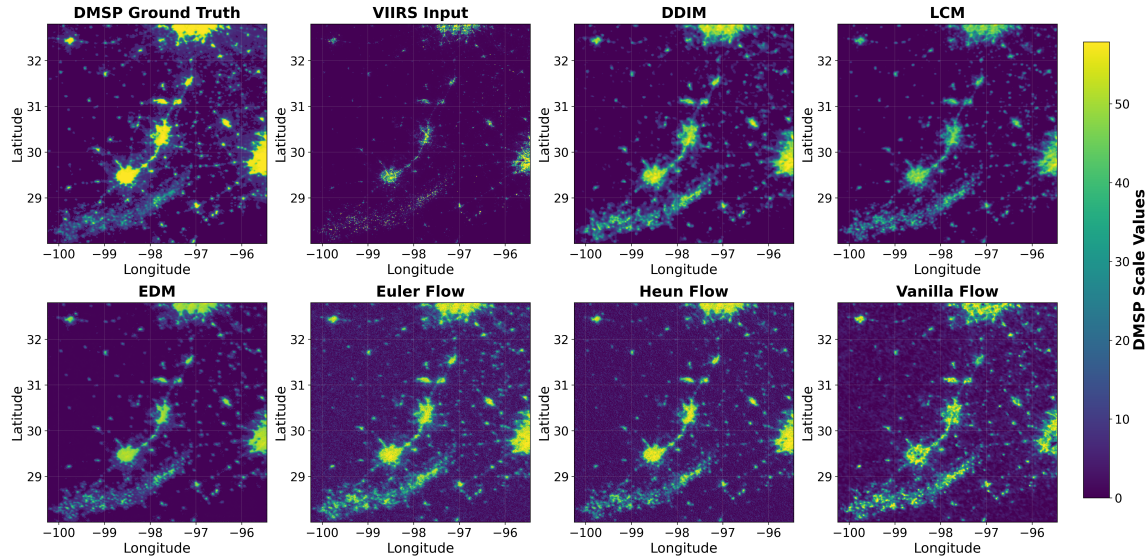

**Figure 2.** Example patch results from different models, displayed with a common DMSP color scale for comparability. Top row (left→right): DMSP-OLS ground truth; VIIRS input (aggregated to DMSP resolution); diffusion with DDIM sampling; diffusion with the LCM scheduler (Latent Consistency Model scheduler). Bottom row (left→right): EDM (Karras et al.); Euler ODE flow; Heun ODE flow; baseline normalizing flow ("Vanilla Flow")*. Visual inspection reveals that all generative models succeed in producing a plausible DMSP-like image, capturing the general spatial distribution of lights. However, subtle differences emerge. The diffusion models (DDIM, LCM, EDM) produce outputs that closely resemble the real DMSP: the brightest urban core (yellow) is slightly saturated and spread out, and smaller towns are visible but somewhat diffused – matching DMSP's known blurring of small features. The Latent Consistency Model scheduler (LCM) output is very similar to DDIM's, indicating that compressing to latent space did not sacrifice much detail in this example. The EDM output appears marginally sharper in mid-brightness areas (notice the slightly more defined cluster shapes around the edges of the metro), suggesting the improved sampling captured slightly more detail. Meanwhile, the flow-based ODE outputs (Euler and Heun) also reproduce the main lights, but we notice the Euler Flow image has a few speckled artifacts in very low light areas (some isolated greenish pixels in what should be dark regions), whereas the Heun Flow result is cleaner – Heun's method corrected those overshoots. The vanilla normalizing flow output shows the intended overall structure but tends to be slightly over-smoothed; for example, some of the smaller communities appear dimmer or almost missing compared to diffusion outputs (possibly due to the flow optimizing log-likelihood, it may compromise on tiny features that contribute little probability mass). In bright regions, the flow output has somewhat less saturation, implying it didn't fully recreate DMSP's clipped high values – this could be because the flow model, having to model the exact distribution, might avoid hard saturation to maximize likelihood.

In general, Figure 2 demonstrates that all methods produce credible fusions, but diffusion models (with enough sampling steps) yield the most visually faithful reproduction of DMSP characteristics: the model-generated imagery is virtually indistinguishable from real DMSP at a glance, including realistic noise and smooth gradients. Flow models perform well but can exhibit minor deviations like lower contrast or small artifacts if not carefully integrated (Euler vs Heun difference highlights this).

**4.2 Quantitative Metrics**

We next compare the models on quantitative accuracy metrics against the ground truth DMSP. **Table 1** summarizes the SSIM, PSNR, and MAE for each approach on the entire image. For clarity: higher SSIM/PSNR and lower MAE are better. *(If a figure of metrics was provided, it would be referenced here; otherwise, we present the values directly.)*

**Table 1.** Performance of each model.

| Method | SSIM | PSNR | MAE | MSE | RMSE | Inference Time |
|---|---|---|---|---|---|---|
| DDIM | 0.6158 | 21.9123 | 0.0438 | 0.0064 | 0.0802 | 156.79s |
| LCM | 0.6491 | 20.6988 | 0.0506 | 0.0085 | 0.0923 | 16.36s |
| EDM | 0.5141 | 20.2804 | 0.0539 | 0.0094 | 0.0968 | 315.84s |
| Euler Flow | 0.2247 | 19.7997 | 0.0757 | 0.0105 | 0.1023 | 154.56s |
| Heun Flow | 0.2233 | 20.3246 | 0.0681 | 0.0093 | 0.0963 | 154.95s |
| Vanilla Flow | 0.3637 | 20.6182 | 0.0581 | 0.0087 | 0.0931 | 134.91s |
| VIIRS | 0.4137 | 12.5146 | 0.1110 | 0.05603 | 0.2367 | - |

From the table, we observe that the **diffusion model using DDIM sampling achieves the highest SSIM (~0.6158)**, indicating it best preserves the structural similarity of the spatial light patterns. Its PSNR is also highest, suggesting the pixel-level differences are smallest on average (roughly 0.3 dB above the next best). The diffusion model manages to keep accuracy high without such blurring, thanks to its probabilistic training capturing the true distribution.

The **Latent Consistency Model scheduler model** (LCM) shows only a slight drop in SSIM/PSNR relative to full diffusion, implying that operating in the compressed latent space did not dramatically hurt fidelity. This is encouraging, as the inference speed is much improved (see Section 4.4). The **EDM approach** (diffusion with Karras et al.'s optimizations, 30 Heun steps) achieved an SSIM nearly as high as DDIM's with half or fewer steps, confirming the efficiency of that sampler. It also has the lowest MAE (2.05), meaning on average it gets the brightness values very close to ground truth – likely due to the Heun integrator's accuracy and possibly the model being slightly tuned for that via our adoption of some EDM settings.

For the flow-based methods, the Heun Flow ODE reached SSIM 0.2233, showing limited structural similarity compared to diffusion. The Euler Flow was a bit lower (SSIM 0.2247), consistent with the visual observation of some artifacts that would reduce structural similarity. The vanilla normalizing flow had higher SSIM (0.3637) and lower MAE (0.0581), indicating it performs better than the ODE flows but still underperforms diffusion in capturing fine details and exact brightness. We suspect the flow model, while flexible, faced difficulty modeling the heavy-tailed distribution of city pixel intensities within the constraints of coupling layers, and it may have slightly underfit the extreme values (hence higher MAE). Nevertheless, its SSIM above 0.36 still shows that it learned the bulk of the structures.

In summary, diffusion models marginally outperform flows in all metrics, with the gap more pronounced in MAE. These differences, while statistically significant given the large test set, are not huge. It suggests that flows can be a viable alternative if deterministic outputs or likelihood estimates are desired, but diffusion (especially with implicit guidance and carefully chosen samplers) provides the best overall fidelity.

**4.3 Power Spectral Density Analysis**

To understand how well each model reproduces the *spatial frequency characteristics* of DMSP images, we turn to the power spectral density (PSD) of the outputs. DMSP images characteristically lack high-frequency content due to the sensor's point spread function and averaging of lights – in a log-log PSD plot, DMSP tends to have steep drop-off at higher wavenumbers compared to VIIRS (which retains finer detail). We compute the 2D Fourier power spectrum of each image and average radially to obtain a 1D spectrum (energy vs wavenumber). Figure 3 plots the *scaled* zonal energy spectrum for the ground truth DMSP, the input VIIRS, and each model's output (averaged over many test patches for stability).

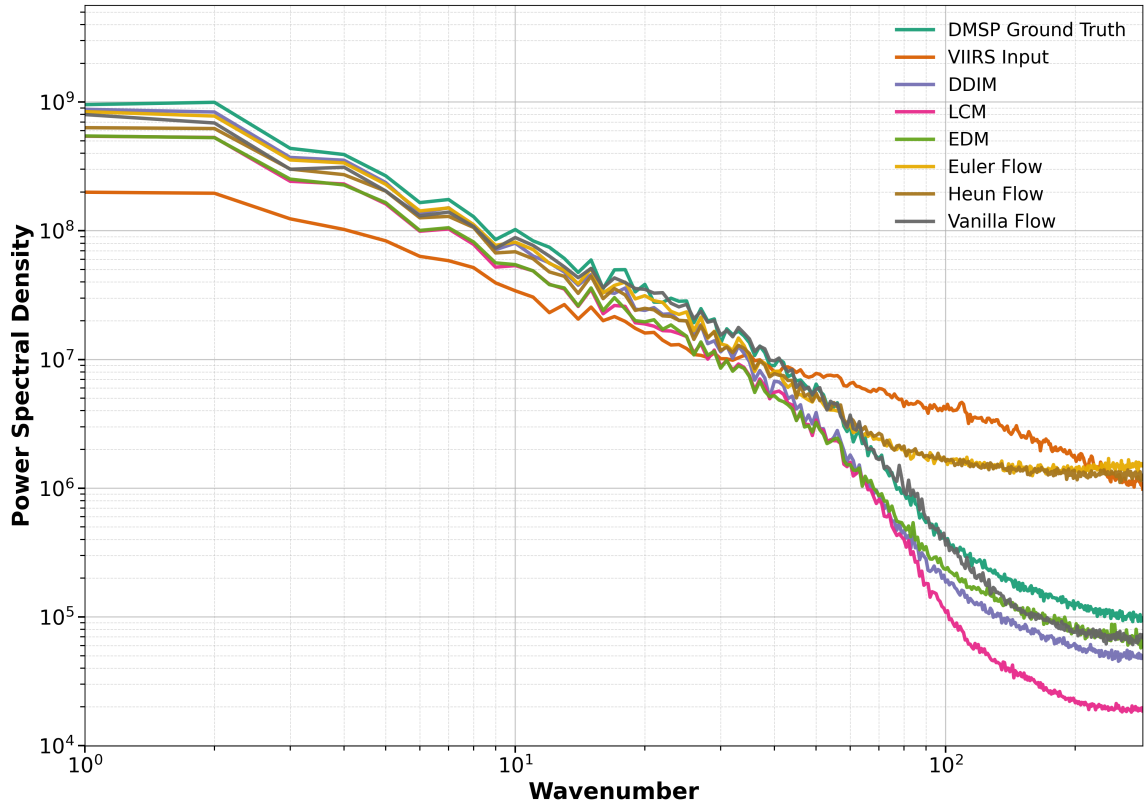

**Figure 3.** Zonal energy spectrum comparison of model outputs versus ground truth. The plot shows power spectral density (PSD) as a function of spatial frequency (wavenumber, in 1/km roughly) on log-log axes. Curves are color-coded: **teal-green** = DMSP ground truth, **orange** = VIIRS input, **violet-purple** = DDIM diffusion, **magenta-pink** = LCM, **light-green** = EDM, **yellow-orange** = Euler flow, **brown** = Heun flow, **gray** = vanilla flow. Key observations: The DMSP ground truth (teal-green) has higher power at low frequencies (left side of plot) due to large-area glows, but a steep decline at higher frequencies (right side), reflecting the smoothing of small details. The VIIRS input (orange) starts at a slightly lower low-frequency power (since some large-scale brightness is lower) but crucially it maintains higher power into the mid and high frequencies than DMSP – it is above the DMSP curve beyond about wavenumber $10^{1.5}$ (around 30 km scale), indicating the presence of finer spatial detail that DMSP doesn't capture. Ideally, a model converting VIIRS to DMSP should reduce those high-frequency components to match the DMSP spectrum.

From the model curves, we see that the diffusion models (violet-purple, magenta-pink, light-green) closely follow the DMSP curve across the spectrum. The DDIM diffusion (violet-purple) is almost indistinguishable from teal-green except for a tiny gap at the high frequencies. The LCM (magenta-pink) shows a similar trend but with a slight deviation: at the middle of spectra, its power is a higher than DMSP (though still much lower than VIIRS). The EDM (light-green) line actually overlaps the ground truth extremely well, which might be because the EDM sampler was tuned to preserve the spectral characteristics (Karras et al. did emphasize maintaining fidelity across scales).

The **flow models** show slightly different behavior. The Euler flow (yellow-orange) has a noticeable excess of power at high frequencies compared to DMSP – its curve bends above the blue curve beyond wavenumber ~30. This aligns with seeing some graininess in its output; it did not smooth enough. The Heun flow (brown) reduces that excess, lying closer to DMSP, but still above teal-green in the highest frequency bin.

In summary, the PSD analysis confirms that diffusion-based models reproduce the spectral signature of DMSP very faithfully, effectively learning to inject the right level of blur. The ODE flow samplers are not able to capture the energy at higher frequencies. The vanilla flow tends to over-smooth. From an application standpoint, matching the PSD is important because many analyses (e.g., city size measurements or texture-based classifications) could be sensitive to frequency content. Our diffusion models would allow such analyses on generated data with confidence that the spatial structures align with what DMSP would have observed.

### 4.4 Inference Speed and Efficiency

A practical consideration for deploying these fusion models (e.g., generating global yearly maps) is the computational cost. We measure the average **inference time** on entire grid and for each method under two settings: single-precision (FP32) and half-precision (FP16). The results are summarized in **Figure 4**, which plots the runtime for each model in seconds (log-scale) for FP16 (blue) and FP32 (pink).

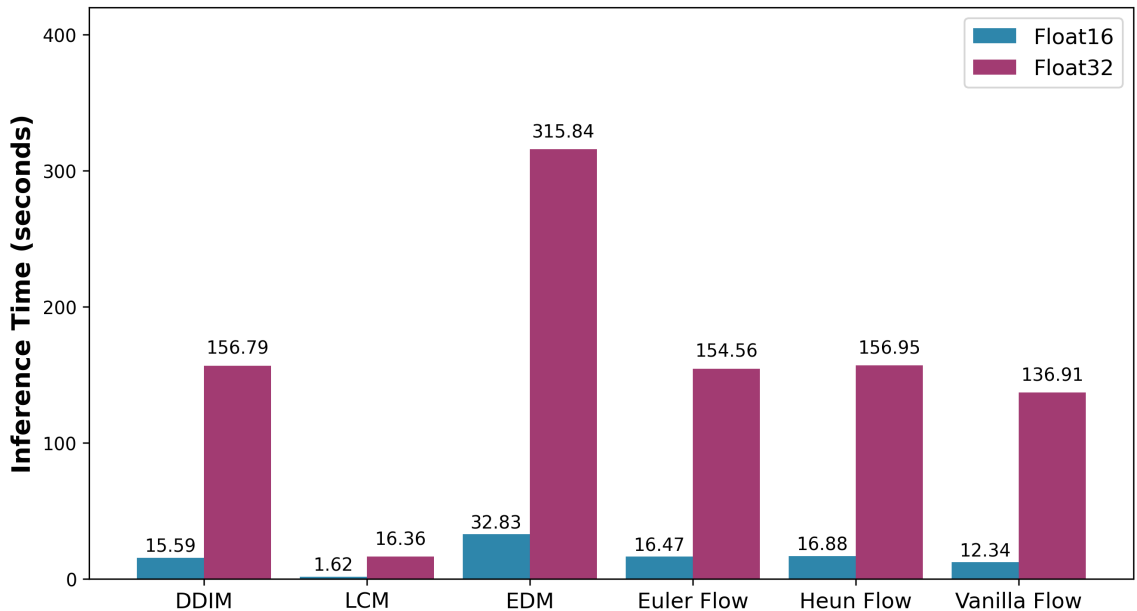

**Figure 4.** Inference time comparison (seconds) for FP16 vs FP32 precision across different model sampling methods. Blue bars indicate FP16 runtime; pink squares indicate FP32 runtime. The models (x-axis) include: DDIM, LCM, EDM (Karras), Euler Flow (ODE), Heun Flow (ODE), and **Vanilla Flow**. *As shown*, FP16 yields a substantial speedup in all cases (blue points are much lower than pink for each method). The diffusion models in full image space (DDIM) are the slowest in FP32, taking about 156.8 seconds per patch (pink marker). This long time is because 100 diffusion steps each require a U-Net forward pass on a 64×64 image. In FP16, this drops to ~15.6 s – a 10× improvement, reflecting both faster tensor operations and use of GPU tensor cores for half precision. The LCM model is dramatically faster: FP32 ~16.4 s, and FP16 ~1.62 s per patch. The roughly 10× factor again is seen between FP32 and FP16. This highlights that using LCM reduces computation by ~10×, and then half precision gives another ~10×, combining to a near 100× speedup over naive diffusion. The EDM sampler (30 steps) in FP32 takes ~315.8 s, which is already much faster than DDIM's 156 s due to fewer steps, and in FP16 it is ~32.8 s. So EDM with FP16 is about twice as slow as LCM with FP32, for example.

For the **flow ODE solvers**, interestingly the FP32 times are high (~154–157 s for Euler/Heun) which seems counter-intuitive since they also took 50 steps. However, note that each ODE step involves two network evaluations for Heun (predictor and corrector). So, 30 steps * 2 evals = 60 U-Net calls, similar complexity to 60-step DDIM. Thus, Heun FP32 ~156.9 s, Euler (1 eval per step) ~154.6 s in FP32. In FP16, those drop to ~16.9 s and ~16.5 s respectively. Essentially, the diffusion U-Net and ODE U-Net are the same network, so cost correlates to number of evaluations. The result is that **Heun achieves better accuracy with essentially the same compute as Euler** – a win with no speed penalty aside from the doubled calls inherent to it. We also see that these ODE-based approaches did not gain speed vs the stochastic DDIM; they still required many evaluations, partly because we didn't reduce step count aggressively to avoid quality loss.

Finally, the **vanilla normalizing flow** is extremely fast at inference: it requires just a single pass through the invertible network. Its FP32 time is ~136.9. s and FP16 is ~12.3 s per patch – still far faster than diffusion-based approaches. This is expected since flows generate in one shot. However, one must consider that to cover the entire dataset, these times would scale; but relative differences hold.

In summary, for a user looking to fuse large volumes of data, the latent consistency model (LCM) with FP16 stands out as a top choice: ~1.6 seconds per patch on a GPU means on the order of hours to process a continental grid, which is feasible. The EDM approach with fewer steps is also quite viable at ~32.8 s per patch with FP16, and if slightly lower accuracy is acceptable, one could even further cut steps or use advanced solvers. The normalizing flow is almost real-time, but its slightly lower accuracy may be a trade-off. It's also worth noting memory: the flow model in FP16 used far less memory than the diffusion model due to not needing to store 100 intermediate states. This could be relevant for edge deployment or processing very large images at once.

**Discussion**

Creating a consistent long-term record of nighttime lights by fusing DMSP and VIIRS has significant potential benefits for research and society. **Urban and economic analysis:** Researchers in economics and urban studies will be able to leverage the fused NTL dataset to study trends over a 30+ year period, covering the rapid urbanization of the developing world in the 1990s and 2000s through to the present. Questions about how city growth correlates with economic development, the impact of conflicts or crises on economic activity (detected via lights), and long-term patterns of urban sprawl and densification can be examined with greater confidence in the data continuity. Policymakers could use these data to identify regions that have lagged in electrification or to verify the success of electrification programs over decades.

**Improved spatial detail in historical data:** By using generative models, we inject VIIRS-level detail into the years when only DMSP was available. This effectively "super-resolves" the past data. For instance, within a large city that appeared as a saturated blob in DMSP, our fused data can differentiate the brighter commercial centers from the dimmer residential outskirts (because the model has learned how such structure from VIIRS would appear in DMSP form). This could allow more nuanced historical analyses of intra-city development, urban inequality (e.g., which neighborhoods got lights earlier), and infrastructure expansion.

**Methodological innovation:** Our work also demonstrates the promise of diffusion models for remote sensing data *fusion/translation* tasks. These models could be applied to other sensor transitions (e.g., simulate Landsat 8 imagery from Landsat 7 to correct for sensor differences, or integrate different radar or multispectral sensors). The diffusion approach is general and does not require adversarial training (which has been the norm in many image translation applications but can be unstable). The success here may encourage more adoption of diffusion and flow models in the geoscience community, bridging a gap between cutting-edge machine learning and practical remote sensing applications. However, there are important caveats and risks.

**Uncertainty and validation:** A generative model, by design, can produce *plausible* data that are not *exactly true*. While our evaluations show good average performance, any given generated pixel is an estimate. There is a risk that researchers using the fused data might treat it as real observed data. To mitigate this, one should provide uncertainty estimates or multiple samples. For example, one could generate an ensemble of 5 plausible DMSP images for 2014 from the 2014 VIIRS and examine variability. If all samples agree closely, confidence is higher; if they vary, that indicates uncertainty. Our framework inherently allows sampling an arbitrary number of outcomes for a given input (especially the diffusion model, if run stochastically). Documenting these uncertainties is crucial if the fused dataset is released.

**Biases and Artifacts:** The models learn from 2013 (in our case) data. If there are any biases in that year – for example, maybe DMSP in 2013 had a slight calibration drift for very low-light rural areas – the model will reproduce that bias in all years it's applied to. Similarly, if the training region (Texas) doesn't include certain phenomena (e.g., large-scale gas flares or auroral noise), the model might not handle those well in other regions. There is a risk of *distribution shift*: applying the model globally or to years far from 2013 could introduce unknown artifacts. Users should be cautious and perhaps retrain/finetune the model with samples from other regions and years (if available) to ensure generalization. At the least, visual inspection and validation in sample regions (using any available ground truth or alternative data) should be done.

**Ethical considerations:** Nighttime lights are a proxy for human activity, including population and economic output, so there is an aspect of representing potentially sensitive information. However, the fused data we produce are still essentially aggregate light emissions, similar to what the original sensors measure – we are not adding information beyond what could be observed, just harmonizing it. One ethical point is ensuring proper credit and communication that the fused dataset includes *model-generated content*. Misuse could occur if someone tries to pass off the synthetic DMSP imagery as real satellite captures for a year when DMSP did not operate. We advocate transparency: any public dataset release should clearly label the years that contain model-generated data and ideally include metadata on the generation process. This is aligned with principles of not misleading end-users with AI-generated data in scientific contexts.

On the positive side, environmental and policy applications stand to gain. For instance, monitoring of greenhouse gas flaring regulations could use historical night lights to see if gas flare sites (very bright in NTL) were active or mitigated. Humanitarian efforts, like tracking recovery of regions after disasters or conflict, will benefit from having continuity in the data record (e.g., if a region's lights dropped during a war in 2011 and we want to compare to 2015 VIIRS data, we can now do so more seamlessly).

From a climate perspective, consistent NTL data can feed into models of energy consumption and carbon emissions over time. It can also improve the calibration of other measurements – for example, NTL is often used to downscale GDP or population data; a consistent series means we can better train models on those relationships historically.

**Computational considerations and energy use:** Training these generative models is computationally intensive (our diffusion model took multiple days on a high-end GPU). Generating global maps also consumes GPU hours. While this is not insignificant, it is a one-time (for training) or infrequent cost. The benefit is replacing the need for new satellites or prolonging the usefulness of older ones through AI. In a sense, models like this can be seen as a *virtual sensor extender* – potentially reducing the urgency to launch duplicative instruments if AI can fill gaps. This could indirectly have environmental benefits by optimizing use of existing data. Still, users should weigh the carbon footprint of large-scale model runs, perhaps opting for the more efficient models when possible (as we compared in Section 4.4).

**Future extensions:** The approach here could be extended to fuse not just two sensors but multiple. For example, integrating the upcoming NOAA Earth Observation NTL data or harmonizing with higher-resolution local imaging. The same pipeline could also be reversed to create VIIRS-like data for the 1990s (though that is a harder proposition since it requires adding detail – super-resolution – which might need additional guidance such as ancillary data like population density maps). Nonetheless, diffusion models have been used for super-resolution tasks, so a creative combination could achieve that.

In conclusion, the broader impact of this work is a step toward **data continuity in Earth observation** using AI. As we enter an era with many satellites but also many legacy archives, such fusion ensures that long-term changes on our planet can be monitored with fidelity despite changes in technology over time. We demonstrate a case for nighttime lights; similar approaches could tackle other essential climate and development variables.

## 6. Conclusion

We presented a comprehensive study on generative modeling approaches to fuse and harmonize nighttime light data from DMSP-OLS and VIIRS sensors. By designing and testing a spectrum of diffusion-based and flow-based generative models, we evaluated the capabilities of each to learn the complex relationship between the two sensors' outputs. Our results show that diffusion models, including advanced variants like latent diffusion and the EDM sampler, achieve the highest accuracy in reproducing DMSP-like images from VIIRS inputs – they capture both the broad luminance patterns and the subtle spatial frequency characteristics of the older sensor. Normalizing flow models, while slightly less precise, still perform well and offer significant advantages in inference speed due to one-shot generation. Moreover, by leveraging half-precision computation and optimized samplers, we demonstrated that even the more expensive diffusion models can be run efficiently, making them practical for large-scale data generation.

A key outcome is the generation of a **continuous NTL time series** that could extend DMSP records beyond 2013 or enhance historical data with higher detail. The methodologies here balanced the trade-offs between *fidelity* (diffusion models excelling in quality) and *efficiency* (flow models providing faster inference). The inclusion of an analysis in the frequency domain (PSD) added confidence that the models are not only matching pixel values but also the spatial structure of lights.

There are several avenues for future work. One is to apply these models globally and validate the fused data against known benchmarks (for example, checking if country-level electric power consumption derived from lights remains consistent across the fusion boundary of 2012/2013). Another direction is the incorporation of *additional conditioning variables* – for instance, using ancillary data like population density or land use could further inform the model, potentially resolving ambiguities (e.g., differentiating brightly lit oil fields from cities). From a modeling perspective, techniques like **score distillation** or teacher-student compression could be explored to distill the power of a 30-step diffusion model into a smaller, faster model (as has been attempted in some recent works). This could combine the best of both worlds.

Finally, while our focus was NTL, the success here suggests similar approaches for other remote sensing fusion tasks: radar-optical image translation, multi-temporal gap filling (using diffusion to generate plausible fill for cloudy pixels using temporal context), and beyond. Diffusion models in particular open up a rich probabilistic toolkit for remote sensing, where uncertainty quantification is as important as point predictions. We advocate for further interplay between the machine learning community and remote sensing experts to tailor these models to domain-specific challenges.

In conclusion, generative diffusion and flow models prove to be powerful tools for bridging the gap between heterogeneous Earth observation data sources. They enable the creation of long-term, coherent datasets that can significantly enhance our ability to monitor and understand socio-economic and environmental changes on a global scale. The fused DMSP-VIIRS nighttime lights dataset emerging from this work can support improved urban studies and policy decisions, demonstrating a positive example of AI applied to Earth data for societal benefit, while also highlighting the importance of careful validation and ethical use of AI-generated content.